\documentclass[conference]{IEEEtran}
\IEEEoverridecommandlockouts
\usepackage{booktabs}
\usepackage{colortbl}
\usepackage[table]{xcolor}

\usepackage{multirow}
\usepackage{cite}
\usepackage{amsmath,amssymb,amsfonts}
\usepackage{algorithmic}
\usepackage{graphicx}
\usepackage{textcomp}
\usepackage{xcolor}
\usepackage{hyperref}

\def\BibTeX{{\rm B\kern-.05em{\sc i\kern-.025em b}\kern-.08em
    T\kern-.1667em\lower.7ex\hbox{E}\kern-.125emX}}
\begin{document}

\definecolor{lightblue}{RGB}{239,239,239} % 浅蓝色
\definecolor{lightgreen}{RGB}{255,243,238} % 浅绿色
\definecolor{lightpink}{RGB}{255,240,240} % 浅粉色

\title{NeuroABench: A Multimodal Evaluation Benchmark for Neurosurgical Anatomy Identification
}

\author{
Ziyang Song\textsuperscript{1, $\dagger$},
Zelin Zang\textsuperscript{1, $\dagger$},
Xiaofan Ye\textsuperscript{2},
Boqiang Xu\textsuperscript{1},
Long Bai\textsuperscript{3},
Jinlin Wu\textsuperscript{1,*},\\
Hongliang Ren\textsuperscript{3},
Hongbin Liu\textsuperscript{1},
Jiebo Luo\textsuperscript{1},
Zhen Lei\textsuperscript{1}

\thanks{This work was supported in part by the National Natural Science Foundation of China (Grant No.\#62306313) and the InnoHK Program by the Hong Kong SAR Government.}
\thanks{\textsuperscript{$\dagger$} Equal contribution.}
\thanks{\textsuperscript{*} Corresponding author.}
\thanks{\textsuperscript{1} Hong Kong Institute of Science and Innovation, Hong Kong SAR, China.}
\thanks{\textsuperscript{2} The University of Hong Kong-Shenzhen Hospital.}
\thanks{\textsuperscript{3} Department of Electronic Engineering, The Chinese University of Hong Kong, Hong Kong SAR, China.}
}

\maketitle

\begin{abstract}
Multimodal Large Language Models (MLLMs) have shown significant potential in surgical video understanding. With improved zero-shot performance and more effective human-machine interaction, they provide a strong foundation for advancing surgical education and assistance. However, existing research and datasets primarily focus on understanding surgical procedures and workflows, while paying limited attention to the critical role of anatomical comprehension. In clinical practice, surgeons rely heavily on precise anatomical understanding to interpret, review, and learn from surgical videos. To fill this gap, we introduce the \textbf{Neuro}surgical \textbf{A}natomy \textbf{Bench}mark (NeuroABench), the first multimodal benchmark explicitly created to evaluate anatomical understanding in the neurosurgical domain. NeuroABench consists of 9 hours of annotated neurosurgical videos covering 89 distinct procedures and is developed using a novel multimodal annotation pipeline with multiple review cycles. The benchmark evaluates the identification of 68 clinical anatomical structures, providing a rigorous and standardized framework for assessing model performance. Experiments on over 10 state-of-the-art MLLMs reveal significant limitations, with the best-performing model achieving only 40.87\% accuracy in anatomical identification tasks. To further evaluate the benchmark, we extract a subset of the dataset and conduct an informative test with four neurosurgical trainees. The results show that the best-performing student achieves 56\% accuracy, with the lowest scores of 28\% and an average score of 46.5\%. While the best MLLM performs comparably to the lowest-scoring student, it still lags significantly behind the group’s average performance. This comparison underscores both the progress of MLLMs in anatomical understanding and the substantial gap that remains in achieving human-level performance. These findings highlight the importance of optimizing MLLMs for neurosurgical applications.
\end{abstract}


\section{Introduction}

In recent years, multimodal large language models (MLLMs) have demonstrated remarkable progress in surgical video understanding, laying a potential foundation for advancing both surgical education and intraoperative assistance~\cite{siuj6010005,bai2023surgical}. Existing studies~\cite{bai2025surgical,chen2024surgfc,hao2025surgery} and datasets~\cite{endonet,wang2025endochat} primarily focus on the recognition of surgical actions, workflows~\cite{chen2024surgplan++,luo2024surgplan}, or tool usage~\cite{li2024llava,wang2024surgical,hao2025surgical}. However, these works tend to overlook the critical aspect of anatomical understanding. In clinical practice, surgeons often rely heavily on the identification and comprehension of anatomical structures to interpret, review, and learn from surgical videos. The lack of datasets and benchmarks centered on anatomical understanding limits the development and evaluation of advanced AI models tailored for real clinical needs.

As shown in Table~\ref{tab:dataset_comparison}, existing medical visual question answering (VQA) datasets pay limited attention to fine-grained anatomical understanding in realistic clinical scenarios. Datasets such as OmniMedVQA~\cite{hu2024omnimedvqanewlargescalecomprehensive}, VQA-RAD~\cite{lau2018dataset}, and SLAKE~\cite{40cb06d16fd1450ea39bfd13d43e9c9f} include some questions related to anatomy, but their focus is typically on broad organ-level identification using static imaging modalities like MRI and CT. These settings do not capture the dynamic, detailed anatomical context crucial for intraoperative operation. While more recent benchmarks, such as GMAI-MMBench~\cite{chen2024gmaimmbenchcomprehensivemultimodalevaluation}, attempt to include anatomical tasks in clinical environments, they still suffer from two key limitations: 1) potential data leakage due to reusing and relabeling from previous datasets, and 2) a lack of comprehensive, fine-grained anatomical classification relevant to surgery. Therefore, currently available multimodal benchmarks fall short of meeting the specific and nuanced needs of surgical AI, particularly for evaluating models' true anatomical comprehension in operative settings.

\begin{figure*}[t]
    \centering
    \includegraphics[width=0.7\linewidth]{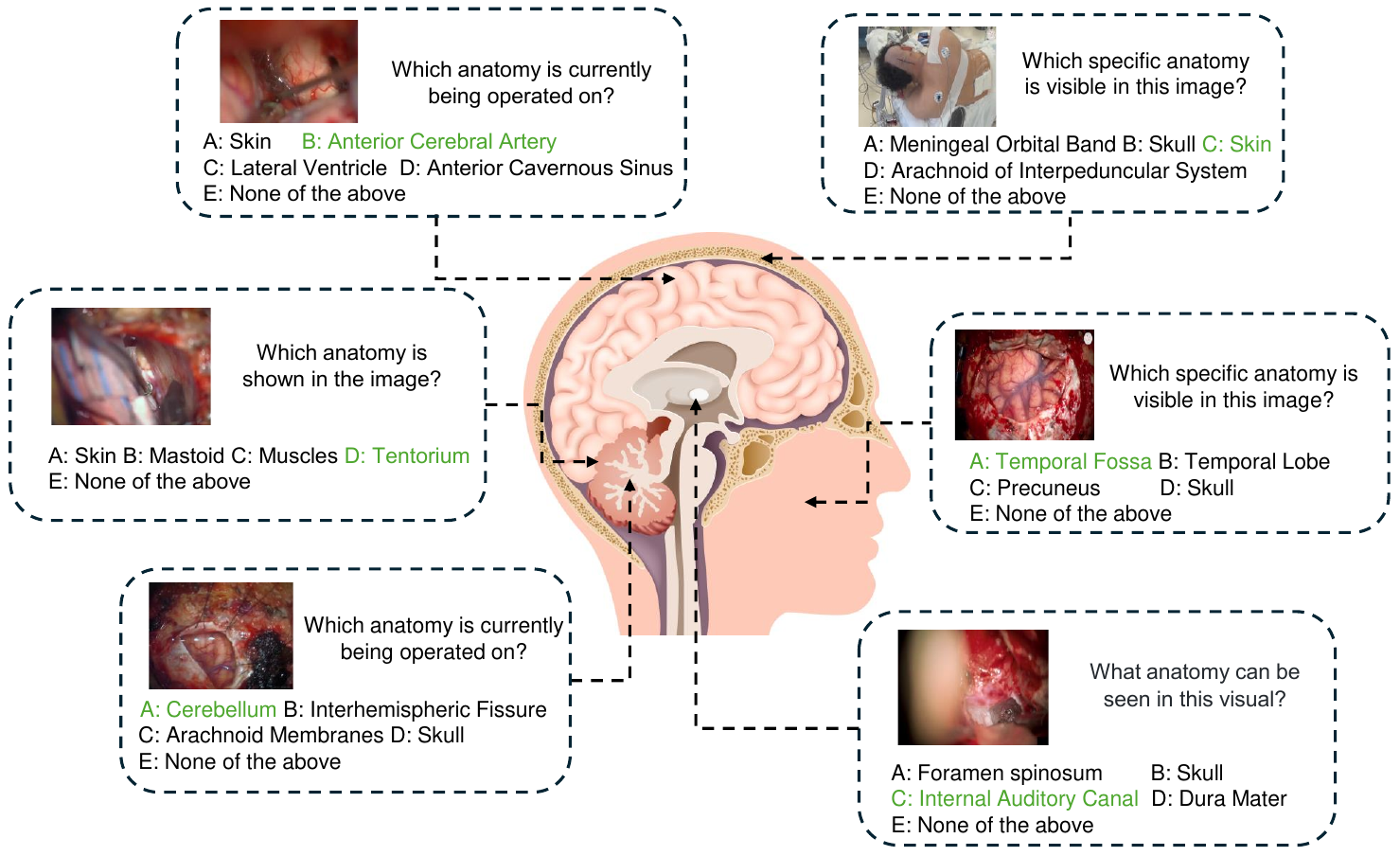}
    \caption{\textbf{Examples of NeuroABench.}  This benchmark is constructed using neurosurgical anatomical content derived from publicly available educational videos. Each video is paired with a question designed to query the identification of specific neuroanatomical structures. For every question, a set of candidate anatomical structures (e.g., skin, anterior cerebral artery, temporal lobe) is provided as multiple-choice options.}
    \label{fig:enter-label}
\end{figure*}

\begin{table*}[]
\centering
\caption{\textbf{Comparison between NeuroABench and other existing medical visual-answer benchmarks. }Here, we only select and count the number related to anatomical understanding from those datasets. *AC denotes the Anatomy Category.}
\label{tab:dataset_comparison}
\begin{tabular}{llcl}
\toprule
\textbf{Dataset}        & \textbf{\textbf{\#}Size} & \textbf{ $\#$AC$^*$} & \textbf{Source} \\ 
\midrule
VQA-RAD         & 106   & 31           & Medpix  \\
SLAKE         & 507   & 21           & MSD, Chestx-ray8, CHAOS\\
GMAI-MMBench    & 2098  & 18           & 284 datasets from public \& hospital       \\
OmniMedVQA & 16448 & 48           & 73 classification datasets       \\
\midrule
NeuroABench(ours)         & 1079  & 68           & Neurosurgical Atlas Website   \\
\bottomrule
\end{tabular}
\end{table*}

Neurosurgical procedures are inherently anatomy-driven: surgeons interpret, review, and learn from surgical videos primarily by recognizing anatomical structures and understanding their relationship to surgical maneuvers. Precise anatomical comprehension underpins intraoperative operation, postoperative assessment, and the continuous improvement of surgical skills. In this context, MLLMs' ability to accurately identify fine-grained anatomical landmarks and associate them with corresponding operative actions is critical for effective surgical video understanding and support. However, despite their strong performance in general multimodal tasks~\cite{Yin_2024}, current MLLMs remain limited in their anatomical reasoning capabilities within neurosurgical settings. They often struggle to capture nuanced anatomical features, spatial relationships, and their dynamic changes during surgery, which are elements essential for surgical quality assessment and skill development~\cite{chen2024asi}. Thus, there is an urgent need for dedicated benchmarks that systematically evaluate and drive the development of MLLMs towards robust anatomical understanding in neurosurgery.

\begin{figure*}[t]
    \centering
    \includegraphics[width=0.9\linewidth]{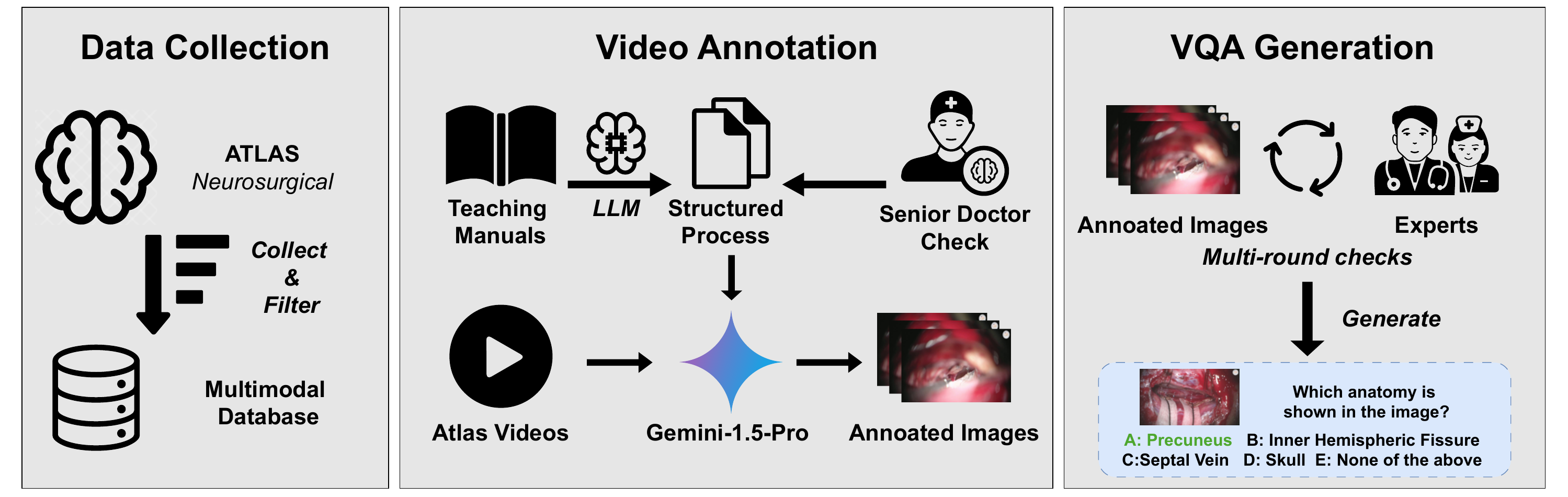}
    \caption{\textbf{Pipeline illustration of NeuroABench.} The data collection can be divided into three main steps: 1) We search hundreds of videos and teaching manuals from the Neurosurgical Atlas, then keep 89 high-quality videos and 32 teaching manuals after filtering. 2) We use Gemini-1.5-Pro to annotate videos with the instructions of clinician-reviewed structured progress extracted from teaching manuals. 3) The annotated images go through additional validation and experts' selection. From these images, we generate question-and-answer pairs for each landmark anatomy featured in the videos.}
    \label{fig:data_build}
\end{figure*}

To address current gaps in anatomical understanding for neurosurgery, we introduce \textbf{NeuroABench} (Figure~\ref{fig:enter-label}), a comprehensive, domain-specific, multimodal benchmark that provides a standardized framework to evaluate MLLMs’ abilities to comprehend anatomical structures and support intraoperative decisions in real-world neurosurgical contexts. NeuroABench builds upon a rigorously curated collection of 89 high-quality neurosurgical videos and 32 clinician-reviewed teaching manuals, selected after filtering hundreds of resources from the Atlas. Using Gemini-1.5-Pro, each video receives annotations with structured progressions extracted from the manuals and validated by expert surgeons; the resulting images undergo further expert review to generate precise question-and-answer pairs for every landmark anatomy depicted. This meticulous pipeline ensures that NeuroABench captures a wide spectrum of clinical approaches and anatomical variations, enabling consistent and reproducible evaluations that foster the development of clinically valuable AI systems for neurosurgery. Building upon this foundation, we conduct comprehensive experiments to highlight the deficiencies of current MLLMs in anatomical recognition. Additionally, we engage four neurosurgical specialist trainees to participate in testing a subset of our dataset, enabling a comparative analysis between the capabilities of state-of-the-art models and those of medical professionals at a general proficiency level.

In summary, our main contributions are as follows: 1) We collect and curate a specialized corpus of neurosurgical videos, comprising 89 approach recordings with a total duration of 9 hours, all annotated and rigorously validated by senior neurosurgeons according to intraoperative guidance standards. 2) We propose \textbf{NeuroABench}, the first benchmark specifically designed to evaluate the anatomical understanding of MLLMs in real-world neurosurgical scenarios, systematically covering 32 neurosurgical approaches and 68 core anatomical structures. 3) Through extensive evaluation of over 10 state-of-the-art MLLMs (spanning both general-purpose and medical-specific MLLMs), we reveal a substantial performance gap between MLLMs and even neurosurgical trainees in neurosurgical anatomical understanding: even the most advanced models achieve only 40.87\% accuracy in anatomical identification tasks, highlighting the urgent need for further research in this domain.

\section{Neurosurigcal Anatomy Bench}
We develop NeuroABench, an innovative benchmark meticulously designed for the neurosurgery field, capable of providing comprehensive evaluations of MLLMs in anatomy identification during the neurosurgical approach. A total of 886 neurosurgical videos are sourced from the Neurosurgical Atlas\footnote[1]{https://www.neurosurgicalatlas.com/}, a globally recognized platform that offers authoritative neurosurgical tutorials presented by expert clinicians. We subsequently filter 89 videos focused on the neurosurgical approach. Based on the data foundation, we design a reliable pipeline to generate question-answer pairs. The steps in constructing our NeuroABench can be divided into three main steps as shown in Figure~\ref{fig:data_build}.

\subsection{Multimodal Data Collection and Curation}
To establish a strong foundation for evaluating the understanding of anatomy, we carefully design the data collection phase. First, we systematically curate video resources from Atlas. Second, to assist with labeling, we acquire neurosurgical approach manuals from Atlas. These manuals provide detailed, step-by-step protocols and critical precautions for various neurosurgical techniques. As a result of this process, we have curated 89 high-quality videos and 32 teaching manuals.

\begin{table*}[t]
    \centering
    \caption{\textbf{Results of the advanced MLLMs on NeuroABench.} The best-performing model in each category is in-bold, and the second best is underlined. $^{*}$DeepSeek-VL2 is based on DeepSeekMoE-27B, with 4.5B parameters activated during inference.}
    \label{tab:performance}
    \begin{tabular}{@{}ll|cccc|ccc@{}}
        \toprule 
        \multirow{2}{*}{Model}            & \multirow{2}{*}{Params} & \multicolumn{4}{c|}{Instance-Level} & \multicolumn{3}{c}{Anatomy-Level}                                                                                                     \\
        \cmidrule(lr){3-6} \cmidrule(lr){7-9}
                                          &                         & Acc                                 & Precision                         & Recall            & F1-Score          & Precision         & Recall            & F1-Score          \\
        \midrule
        Random                            & -                       & 19.48                               & 19.99                             & 20.01             & 19.96             & 16.86              & 12.10              & 1.68              \\
        \addlinespace
        \rowcolor{lightblue}
        \multicolumn{9}{c}{\textbf{Open-Source MLLMs}}                                                                                                                                                                                            \\
        \addlinespace
        mPLUG-Owl3 \cite{ye2024mplugowl3longimagesequenceunderstanding}                      & 8B                      & 25.76                               & 20.37                             & 25.36             & 22.55             & 19.53             & 17.96             & 15.80             \\
        Deepseek-VL2$^{*}$\cite{wu2024deepseekvl2mixtureofexpertsvisionlanguagemodels}                & 4.5B                    & 22.15                               & 27.54                             & 23.35             & 16.93             & 17.53             & 14.96             & 14.65             \\
        LLaVA-NeXT \cite{liu2024llavanext}                       & 7B                      & 28.82                               & 24.06                             & 28.90             & 25.67             & 23.75             & 18.82             & 18.64             \\
        Baichuan-Omni-1.5\cite{li2025baichuan}                 & 7B                      & 27.25                               & 27.90                             & 27.22             & 27.29             & 20.45             & 15.07             & 15.10             \\
        Qwen2.5-VL\cite{Qwen2.5-VL}                        & 7B                      & 34.11                               & 36.21                             & 33.69             & 32.18             & 21.00             & 20.16             & 18.31             \\
        LLaVA-NeXT\cite{liu2024llavanext}                        & 72B                     & 27.53                               & 26.70                             & 27.43             & 25.45             & 16.72             & 16.62             & 14.60             \\
        Qwen2.5-VL \cite{Qwen2.5-VL}                       & 72B                     & 37.44                               & 40.57                             & 37.27             & 37.52             & 25.91             & 20.20             & 20.17             \\
        \addlinespace
        \rowcolor{lightgreen}
        \multicolumn{9}{c}{\textbf{Proprietary MLLMs}}                                                                                                                                                                                            \\
        \addlinespace
        GPT-4o \cite{openai2024gpt4ocard}                          & -                       & 30.21                               & 38.19                             & 29.59             & 29.93             & 25.11             & 20.48             & 19.56             \\
        Gemini-1.5-Pro                    & -                       & 38.83                               & 41.68                             & 38.92             & 35.88             & 22.54             & 23.35             & 20.57             \\
        Gemini-2.0-Flash \cite{geminiteam2024gemini15unlockingmultimodal}                 & -                       & \textbf{40.87}                      & \textbf{46.61}                    & \textbf{41.07}    & \textbf{38.56}    & \textbf{29.68} & \underline{27.02} & \textbf{25.52}    \\
        Qwen-VL-MAX      \cite{qwen1.5}                 & -                       & 18.26                               & 19.46                             & 18.26             & 18.31             & 12.13             & 11.52             & 9.29              \\
        Claude-3.5-Sonnet~\cite{claude3.5_2025}                 & -                       & \underline{40.41}                   & \underline{42.52}                 & \underline{40.44} & \underline{38.12} & \underline{29.44}    & \textbf{27.03}    & \underline{24.52} \\
        Claude-3.7-Sonnet\cite{claude3.7_2025}                 & -                       & 33.27                               & 37.63                             & 33.11             & 30.98             & 25.29             & 20.00             & 20.35             \\
        \addlinespace
        \rowcolor{lightpink}
        \multicolumn{9}{c}{\textbf{Medical Special Models}}                                                                                                                                                                                       \\
        \addlinespace
        LLaVA-Med-v1.5\cite{liu2023llava}                   & 7B                      & 15.01                               & 12.54                             & 15.09             & 7.66              & 13.89             & 10.53             & 10.79             \\
        HuatuoGPT-Vision                  & 7B                      & 30.31                               & 34.28                             & 30.87             & 28.31             & 22.26             & 20.98             & 18.85             \\
        HuatuoGPT-Vision\cite{chen2024huatuogptvisioninjectingmedicalvisual}                  & 34B                     & 36.52                               & 36.40                             & 35.48             & 32.71             & 24.17             & 23.76             & 21.49             \\
        \bottomrule
    \end{tabular}
\end{table*}

\subsection{Video Annotation and Labeling Pipeline}
In this stage, we design a multi-modal pipeline to annotate the videos. First, OpenAI-o1\cite{openai_dynamic_2025} is employed to analyze the neurosurgical approach manuals obtained from ATLAS, automatically summarizing procedural workflows based on anatomical landmarks. These summaries are then reviewed and corrected by a neurosurgical expert to ensure clinical validity and terminological precision. Subsequently, we employ Gemini-1.5-Pro\cite{geminiteam2024gemini15unlockingmultimodal} to annotate pre-selected high-quality neurosurgical videos. During annotation, Gemini-1.5-Pro utilizes both visual content and speech from experienced surgeons within the videos to refine understanding. For each video, Gemini-1.5-Pro generates a JSON file containing three structured keys for each video: (1) ``step number" indicating the sequential order of procedural phases, (2) ``key anatomy" specifying the landmark anatomy identified in each phase, and (3) ``time period" documenting the temporal boundaries of critical procedural steps through timestamp annotations.

For label standardization, our standardization approach has two primary objectives: First, we expand medical abbreviations into their full clinical terms, such as converting ``IAC" to ``Internal Auditory Canal" for better clarity. Second, we aim to unify synonymous terms for anatomy, exemplified by standardizing variations like ``Dura" and ``Dura Mater" into the unified term ``Dura Mater." By implementing these two strategies, terminological expansion and synonym unification, we achieve consistency of the label that highlights crucial medical semantics while reducing interpretive biases caused by variations in vocabulary.

\subsection{QA Generation and Selection}
Based on the annotated videos, we extract frames at one-second intervals and align them with anatomy annotations. Next, we organize a multi-round examination process involving clinical doctors to verify these image-anatomy pairs. During quality control, we remove three types of cases: (1) image-anatomy pair mismatches where annotations contradict the visual evidence; (2) low-quality images lacking clear anatomical details; and (3) frames containing multiple landmark anatomies that could introduce ambiguity.

For question formulation, we predefine a curated question pool consisting of five distinct questions. For each image-anatomy pair, one question is randomly selected from this pool for each case. The answer options follow a standardized format: options A-D present specific anatomical names, while option E serves as a fallback choice (``None of the above"). To maintain a balanced answer distribution, we implement an equal probability allocation strategy where each option has an identical 20\% chance of being designated as the correct answer. As a result, we finalize \textbf{1079} QA pairs for the NeuroABench.

\section{Experiment}
\subsection{Experiment Setup}
In this study, we conduct comprehensive evaluations of diverse MLLMs spanning medical-specialized, open-source, and proprietary API-based general models. The assessment employs a zero-shot paradigm. To evaluate the model’s performance, we use macro-averaged Precision, Recall, and F1-score as the evaluation metrics, similar to EndoNet~\cite{7519080}. If a model’s output does not include clearly followed instructions to select an answer or letter options, it is treated as an error. To address the potential influence of uneven data distribution in our benchmark, we report performance scores at two distinct levels: \textbf{the instance-level}, which represents the average of the performance metrics computed across all individual images, and \textbf{the anatomy-level}, which is derived as the average of the performance metrics across all anatomical structures.

\begin{figure*}
    \centering
    \includegraphics[width=0.6\linewidth]{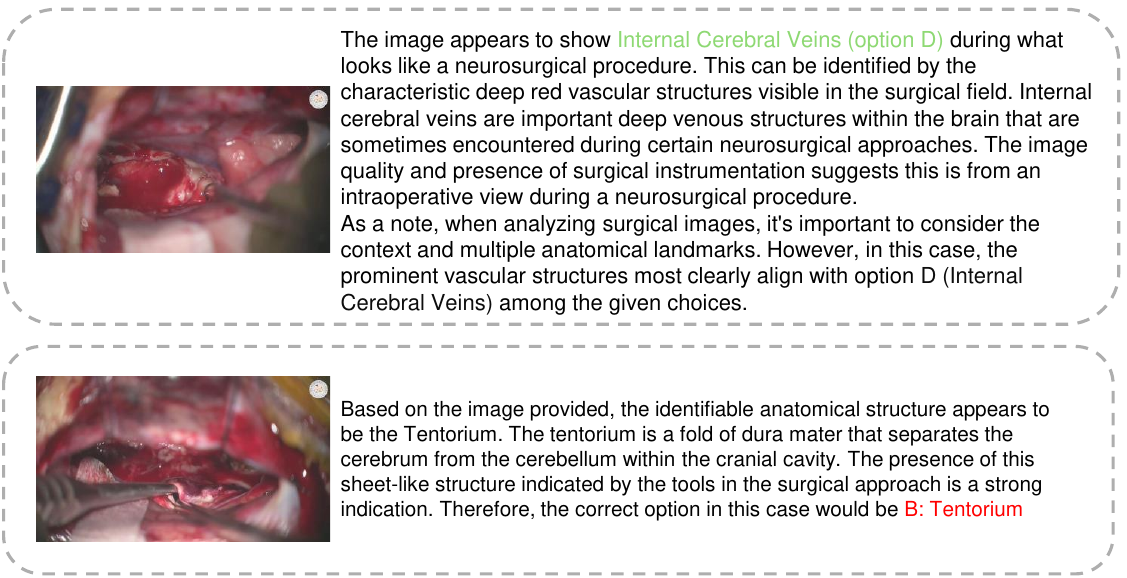}
    \caption{\textbf{A case on the influence of anatomical deformation.} Here, we select the responses of Claude-3.5-Sonnet to two closely similar frames of anatomical images from the same surgical procedure to demonstrate the impact of anatomical deformation on the model's anatomical recognition.}
    \label{fig:case1}
\end{figure*}

\subsection{Models}
For completeness, we conduct evaluations using several state-of-the-art MLLMs to benchmark their performance on SABench, including both general models and medical-specific models that are meticulously trained for clinical medicine. Our testing encompasses architectures ranging from 7 billion to 72 billion parameters to examine scale-related performance variations. Model weights are rigorously sourced from their official Hugging Face repositories, maintaining version integrity across all implementations. The assessment employs a strict zero-shot paradigm where evaluation prompts contain no exemplar demonstrations, requiring models to solve novel problems through pure comprehension without task-specific training or contextual examples. For medical-specific models, we include 3 latest powerful MLLMs: LLaVA-Med-v1.5~\cite{liu2023llava}, HuatuoGPT-Vision-7B, and HuatuoGPT-Vision-34B~\cite{chen2024huatuogptvisioninjectingmedicalvisual}. For general models, we test 7 of the latest and most advanced MLLMs currently available: mPLUG-Owl3~\cite{ye2024mplugowl3longimagesequenceunderstanding}, Deepseek-VL2~\cite{wu2024deepseekvl2mixtureofexpertsvisionlanguagemodels}, LLaVA-NeXT-7B, LLaVA-NeXT-72B~\cite{liu2024llavanext}, Baichuan-Omni-1.5~\cite{li2025baichuan}, Qwen2.5-VL-7B, Qwen2.5-VL-72B~\cite{Qwen2.5-VL}. In addition, we also evaluate 6 proprietary MLLMs via API: GPT-4o~\cite{openai2024gpt4ocard}, Gemini-1.5-pro~\cite{geminiteam2024gemini15unlockingmultimodal}, Gemini-2.0-Flash~\cite{gemini2.0pro_2025}, Qwen-VL-MAX~\cite{qwen1.5}, Claude-3.5-Sonnet~\cite{claude3.5_2025}, and Claude-3.7-Sonnet~\cite{claude3.7_2025}. We also invite four medical trainees to participate in our benchmark assessment.

\subsection{Results}
\begin{table}[htbp]
  \centering
  \caption{\textbf{Performance Comparison of Models and Neurosurgical Specialist Trainees.} We select the best-performing Gemini-2.0-flash for comparison with humans (trainees).}
  \begin{tabular}{lcc}
    \toprule
    \textbf{Participant} & \textbf{Acc} \\ 
    \midrule
    Neurosurgical Specialist Trainees 1 & 0.48 \\ 
    Neurosurgical Specialist Trainees 2 & 0.28 \\ 
    Neurosurgical Specialist Trainees 3 & 0.56 \\ 
    Neurosurgical Specialist Trainees 4 & 0.54 \\ 
    \midrule
    Gemini-2.0-Flash & 0.36 \\ 
    \bottomrule
  \end{tabular}
  \label{tab:performance_comparison}
\end{table}

After reviewing the evaluation results, we have 3 key findings for developing future MLLMs in the medical domain:
\\
\textbf{Finding 1: Medical tasks remain challenging for MLLMs.} Our surgical anatomy identification benchmark, NeuroABench, presents deceptively simple yet clinically demanding tasks. The evaluation reveals that even state-of-the-art models, such as Gemini-2.0-Flash, achieve only 40.87\% precision (Table \ref{tab:performance}), falling short of clinical applicability standards. This performance gap underscores the need for substantial improvements in current MLLMs' medical capabilities.
\\
\textbf{Finding 2: The most advanced models still lag medical trainees.} As shown in Table \ref{tab:performance_comparison}, the accuracies of the four participants are 48\%, 28\%, 56\%, and 54\%, respectively. This indicates that most models can reach the level of the worst medical trainee. However, the best-performing model, Gemini-2.0-Flash, still lags behind most medical trainees by a significant margin (average -10.50\%). Additionally, through our testing, the performance variance of MLLMs is far lower than the variability observed among humans. This demonstrates that while current MLLMs still underperform compared to human practitioners, their future development could enable them to achieve comparable proficiency to doctors while maintaining stability in clinical applications.
\\
\textbf{Finding 3: There is a deficiency in current medical training data.} Our analysis shows that while medically-specialized models (e.g., Huatuo-Vision-34B) achieve modest performance improvements (average +7.52\% accuracy) over general-purpose MLLMs in anatomy reasoning tasks, their absolute performance remains unsatisfying (ceiling accuracy: 36.52\% in Table \ref{tab:performance}). We suspect that current medical special models neglect clinical surgical anatomy comprehension when training with medical data. 

\begin{equation*}
\text{Instance-Level Metric} = \frac{1}{N} \sum_{i=1}^{N} \left( \frac{1}{K} \sum_{j=1}^{K} M_{i,j} \right)
\end{equation*}

\subsection{Discussion}
To gain deeper insight into the reasoning processes of MLLMs, we require these models to produce free-form responses and explicitly articulate their reasoning steps, rather than merely selecting from a set of predefined options. This approach facilitates a more nuanced and comprehensive analysis of how models arrive at their conclusions, allowing us to dissect their decision-making paths and uncover specific error patterns. By examining the explanations provided in these open-ended responses, we can better identify the limitations and strengths of current models in complex clinical scenarios.

The task of clinical anatomical identification presents significant challenges for MLLMs, stemming not only from their limited exposure to detailed anatomical information during pre-training but also from the complex morphological changes induced by intraoperative procedures. These challenges are evident when anatomical structures undergo transformation due to manipulation or surgical intervention, which frequently occurs in real-world neurosurgical settings. For instance, as illustrated in Figure \ref{fig:case1}, in the first image, Claude-3.5-sonnet was able to accurately identify and describe the anatomical features presented, as the anatomy was in the exposure phase and retained its standard morphology, unaffected by surgical manipulation. However, in the second image, where the anatomical structure was actively manipulated and had experienced noticeable deformation, Claude-3.5-sonnet's description deviated from the correct answer, leading to an erroneous final response. This example highlights how morphological alterations can significantly impact model performance.

Such findings underscore the fact that current MLLMs possess an understanding of anatomy that is largely limited to static, textbook representations, rather than the dynamic and often altered presentations encountered in clinical practice. As a result, these models struggle to generalize their anatomical knowledge to the complexities of real-world surgical environments, demonstrating the urgent need for more robust and context-aware training paradigms that can bridge this critical gap between theoretical knowledge and practical application in neurosurgery.

\section{Conclusion}
We collect a diverse set of neurosurgical videos from the Neurosurgical Atlas and establish a systematic pipeline to annotate this new video library. Building on this rigorously curated database, we develop NeuroABench, a benchmark for evaluating neurosurgical anatomy identification by AI models. Our annotation process is overseen by experienced neurosurgeons to ensure clinical accuracy and reliability. Through comprehensive experiments with over 10 advanced MLLMs, we observe notable performance gaps between the MLLMs and neurological trainees. The highest anatomical identification accuracy achieved is only 40.87\%, highlighting the limitations of existing models and emphasizing the need for AI solutions tailored to the unique demands of neurosurgical practice. NeuroABench helps address this gap by offering a standardized, clinically aligned benchmark for evaluating and guiding the development of MLLMs in neurosurgery. 

\bibliographystyle{IEEEtran}
\bibliography{mybibliography}

@article{chen2024huatuogptvisioninjectingmedicalvisual,
  title={Huatuogpt-vision, towards injecting medical visual knowledge into multimodal llms at scale},
  author={Chen, Junying and Gui, Chi and Ouyang, Ruyi and Gao, Anningzhe and Chen, Shunian and Chen, Guiming Hardy and Wang, Xidong and Zhang, Ruifei and Cai, Zhenyang and Ji, Ke and others},
  journal={arXiv preprint arXiv:2406.19280},
  year={2024}
}

@inproceedings{chen2024asi,
  title={Asi-seg: Audio-driven surgical instrument segmentation with surgeon intention understanding},
  author={Chen, Zhen and Zhang, Zongming and Guo, Wenwu and Luo, Xingjian and Bai, Long and Wu, Jinlin and Ren, Hongliang and Liu, Hongbin},
  booktitle={2024 IEEE/RSJ International Conference on Intelligent Robots and Systems (IROS)},
  pages={13773--13779},
  year={2024},
  organization={IEEE}
}

@article{ye2024mplugowl3longimagesequenceunderstanding,
  title={mplug-owl3: Towards long image-sequence understanding in multi-modal large language models},
  author={Ye, Jiabo and Xu, Haiyang and Liu, Haowei and Hu, Anwen and Yan, Ming and Qian, Qi and Zhang, Ji and Huang, Fei and Zhou, Jingren},
  journal={arXiv preprint arXiv:2408.04840},
  year={2024}
}

@article{wu2024deepseekvl2mixtureofexpertsvisionlanguagemodels,
  title={Deepseek-vl2: Mixture-of-experts vision-language models for advanced multimodal understanding},
  author={Wu, Zhiyu and Chen, Xiaokang and Pan, Zizheng and Liu, Xingchao and Liu, Wen and Dai, Damai and Gao, Huazuo and Ma, Yiyang and Wu, Chengyue and Wang, Bingxuan and others},
  journal={arXiv preprint arXiv:2412.10302},
  year={2024}
}

@article{li2024llava,
  title={Llava-surg: towards multimodal surgical assistant via structured surgical video learning},
  author={Li, Jiajie and Skinner, Garrett and Yang, Gene and Quaranto, Brian R and Schwaitzberg, Steven D and Kim, Peter CW and Xiong, Jinjun},
  journal={arXiv preprint arXiv:2408.07981},
  year={2024}
}

@inproceedings{hao2025surgical,
  title={Surgical-MambaLLM: Mamba2-Enhanced Multimodal Large Language Model for VQLA in Robotic Surgery},
  author={Hao, Pengfei and Wang, Hongqiu and Li, Shuaibo and Xing, Zhaohu and Yang, Guang and Wu, Kaishun and Zhu, Lei},
  booktitle={International Conference on Medical Image Computing and Computer-Assisted Intervention},
  pages={573--583},
  year={2025},
  organization={Springer}
}

@misc{liu2024llavanext,
    title={LLaVA-NeXT: Improved reasoning, OCR, and world knowledge},
    url={https://llava-vl.github.io/blog/2024-01-30-llava-next/},
    author={Liu, Haotian and Li, Chunyuan and Li, Yuheng and Li, Bo and Zhang, Yuanhan and Shen, Sheng and Lee, Yong Jae},
    month={January},
    year={2024}
}

@article{hao2025surgery,
  title={Surgery-R1: Advancing Surgical-VQLA with Reasoning Multimodal Large Language Model via Reinforcement Learning},
  author={Hao, Pengfei and Li, Shuaibo and Wang, Hongqiu and Kou, Zhizhuo and Zhang, Junhang and Yang, Guang and Zhu, Lei},
  journal={arXiv preprint arXiv:2506.19469},
  year={2025}
}

@article{liu2023llava,
  title={Visual instruction tuning},
  author={Liu, Haotian and Li, Chunyuan and Wu, Qingyang and Lee, Yong Jae},
  journal={Advances in neural information processing systems},
  volume={36},
  pages={34892--34916},
  year={2023}
}

@article{li2025baichuan,
  title={Baichuan-Omni-1.5 Technical Report},
  author={Li, Yadong and Liu, Jun and Zhang, Tao and Chen, Song and Li, Tianpeng and Li, Zehuan and Liu, Lijun and Ming, Lingfeng and Dong, Guosheng and Pan, Da and others},
  journal={arXiv preprint arXiv:2501.15368},
  year={2025}
}

@misc{Qwen2.5-VL,
    title = {Qwen2.5-VL},
    url = {https://qwenlm.github.io/blog/qwen2.5-vl/},
    author = {Qwen Team},
    month = {January},
    year = {2025}
}

@article{openai2024gpt4ocard,
  title={Gpt-4o system card},
  author={Hurst, Aaron and Lerer, Adam and Goucher, Adam P and Perelman, Adam and Ramesh, Aditya and Clark, Aidan and Ostrow, AJ and Welihinda, Akila and Hayes, Alan and Radford, Alec and others},
  journal={arXiv preprint arXiv:2410.21276},
  year={2024}
}

@misc{qwen1.5,
    title = {Introducing Qwen1.5},
    url = {https://qwenlm.github.io/blog/qwen1.5/},
    author = {Qwen Team},
    month = {February},
    year = {2024}
}

@inproceedings{bai2023surgical,
  title={Surgical-VQLA: Transformer with gated vision-language embedding for visual question localized-answering in robotic surgery},
  author={Bai, Long and Islam, Mobarakol and Seenivasan, Lalithkumar and Ren, Hongliang},
  booktitle={2023 IEEE International Conference on Robotics and Automation (ICRA)},
  pages={6859--6865},
  year={2023},
  organization={IEEE}
}

@article{geminiteam2024gemini15unlockingmultimodal,
  title={Gemini 1.5: Unlocking multimodal understanding across millions of tokens of context},
  author={Team, Gemini and Georgiev, Petko and Lei, Ving Ian and Burnell, Ryan and Bai, Libin and Gulati, Anmol and Tanzer, Garrett and Vincent, Damien and Pan, Zhufeng and Wang, Shibo and others},
  journal={arXiv preprint arXiv:2403.05530},
  year={2024}
}

@article{wang2025endochat,
  title={EndoChat: Grounded Multimodal Large Language Model for Endoscopic Surgery},
  author={Wang, Guankun and Bai, Long and Wang, Junyi and Yuan, Kun and Li, Zhen and Jiang, Tianxu and He, Xiting and Wu, Jinlin and Chen, Zhen and Lei, Zhen and others},
  journal={arXiv preprint arXiv:2501.11347},
  year={2025}
}

@misc{openai_dynamic_2025,
  title        = {OpenAI Model Versioning System},
  author       = {OpenAI},
  year         = {2025},
  url          = {https://platform.openai.com/docs/model-versioning},
  note         = {Model updates occur every 3 months with backward compatibility}
}

@misc{gemini2.0pro_2025,
  title        = {Gemini 2.0 Pro Model Documentation},
  author       = {{Google DeepMind}},
  year         = {2025},
  howpublished = {\url{https://ai.google.dev/gemini-api/docs/models/gemini}},
  note         = {Accessed: 2025-02-19},
  version      = {Experimental (gemini-2.0-pro-exp-02-05)},
  url-date     = {2025-02}
}

@misc{claude3.5_2025,
  title        = {Claude 3.5 Sonnet Model Documentation},
  author       = {Anthropic Team},
  year         = {2024},
  howpublished = {\url{https://www.anthropic.com/news/claude-3-5-sonnet}},
  note         = {Accessed: 2024-06-21}
}

@misc{claude3.7_2025,
  title        = {Claude 3.7 Sonnet Model Documentation},
  author       = {Anthropic Team},
  year         = {2025},
  howpublished = {\url{https://www.anthropic.com/news/claude-3-7-sonnet}},
  note         = {Accessed: 2025-02-25}
}

@article{lau2018dataset,
    title={A dataset of clinically generated visual questions and answers about radiology images},
    author={Lau, Jason J and Gayen, Soumya and Ben Abacha, Asma and Demner-Fushman, Dina},
    journal={Scientific data},
    volume={5},
    number={1},
    pages={1--10},
    year={2018},
    publisher={Nature Publishing Group}
}

@inproceedings{40cb06d16fd1450ea39bfd13d43e9c9f,
  title={Slake: A semantically-labeled knowledge-enhanced dataset for medical visual question answering},
  author={Liu, Bo and Zhan, Li-Ming and Xu, Li and Ma, Lin and Yang, Yan and Wu, Xiao-Ming},
  booktitle={2021 IEEE 18th international symposium on biomedical imaging (ISBI)},
  pages={1650--1654},
  year={2021},
  organization={IEEE}
}

@article{chen2024gmaimmbenchcomprehensivemultimodalevaluation,
  title={Gmai-mmbench: A comprehensive multimodal evaluation benchmark towards general medical ai},
  author={Ye, Jin and Wang, Guoan and Li, Yanjun and Deng, Zhongying and Li, Wei and Li, Tianbin and Duan, Haodong and Huang, Ziyan and Su, Yanzhou and Wang, Benyou and others},
  journal={Advances in Neural Information Processing Systems},
  volume={37},
  pages={94327--94427},
  year={2024}
}

@inproceedings{hu2024omnimedvqanewlargescalecomprehensive,
  title={Omnimedvqa: A new large-scale comprehensive evaluation benchmark for medical lvlm},
  author={Hu, Yutao and Li, Tianbin and Lu, Quanfeng and Shao, Wenqi and He, Junjun and Qiao, Yu and Luo, Ping},
  booktitle={Proceedings of the IEEE/CVF Conference on Computer Vision and Pattern Recognition},
  pages={22170--22183},
  year={2024}
}

@article{7519080,
  title={Endonet: a deep architecture for recognition tasks on laparoscopic videos},
  author={Twinanda, Andru P and Shehata, Sherif and Mutter, Didier and Marescaux, Jacques and De Mathelin, Michel and Padoy, Nicolas},
  journal={IEEE transactions on medical imaging},
  volume={36},
  number={1},
  pages={86--97},
  year={2016},
  publisher={IEEE}
}

@article{siuj6010005,
  title={A Review of the Intraoperative Use of Artificial Intelligence in Urologic Surgery},
  author={Guduguntla, Arjun and Al-Khanaty, Abdullah and Davey, Catherine E and Patel, Oneel and Ta, Anthony and Ischia, Joseph},
  journal={Soci{\'e}t{\'e} Internationale d’Urologie Journal},
  volume={6},
  number={1},
  pages={5},
  year={2025},
  publisher={MDPI}
}

@article{wang2024surgical,
  title={Surgical-lvlm: Learning to adapt large vision-language model for grounded visual question answering in robotic surgery},
  author={Wang, Guankun and Bai, Long and Nah, Wan Jun and Wang, Jie and Zhang, Zhaoxi and Chen, Zhen and Wu, Jinlin and Islam, Mobarakol and Liu, Hongbin and Ren, Hongliang},
  journal={arXiv preprint arXiv:2405.10948},
  year={2024}
}

@article{bai2025surgical,
  title={Surgical-VQLA++: Adversarial contrastive learning for calibrated robust visual question-localized answering in robotic surgery},
  author={Bai, Long and Wang, Guankun and Islam, Mobarakol and Seenivasan, Lalithkumar and Wang, An and Ren, Hongliang},
  journal={Information Fusion},
  volume={113},
  pages={102602},
  year={2025},
  publisher={Elsevier}
}

@article{Yin_2024,
   title={A survey on multimodal large language models},
   volume={11},
   ISSN={2053-714X},
   number={12},
   journal={National Science Review},
   publisher={Oxford University Press (OUP)},
   author={Yin, Shukang and Fu, Chaoyou and Zhao, Sirui and Li, Ke and Sun, Xing and Xu, Tong and Chen, Enhong},
   year={2024},
   month=nov 
}

@inproceedings{chen2024surgfc,
  title={Surgfc: Multimodal surgical function calling framework on the demand of surgeons},
  author={Chen, Zhen and Luo, Xingjian and Wu, Jinlin and Chan, Danny TM and Lei, Zhen and Ourselin, Sebastien and Liu, Hongbin},
  booktitle={2024 IEEE International Conference on Bioinformatics and Biomedicine (BIBM)},
  pages={3076--3081},
  year={2024},
  organization={IEEE}
}

@article{chen2024surgplan++,
  title={SurgPLAN++: Universal Surgical Phase Localization Network for Online and Offline Inference},
  author={Chen, Zhen and Luo, Xingjian and Wu, Jinlin and Bai, Long and Lei, Zhen and Ren, Hongliang and Ourselin, Sebastien and Liu, Hongbin},
  journal={arXiv preprint arXiv:2409.12467},
  year={2024}
}

@inproceedings{luo2024surgplan,
  title={Surgplan: Surgical phase localization network for phase recognition},
  author={Luo, Xingjian and Pang, You and Chen, Zhen and Wu, Jinlin and Zhang, Zongmin and Lei, Zhen and Liu, Hongbin},
  booktitle={2024 IEEE International Symposium on Biomedical Imaging (ISBI)},
  pages={1--5},
  year={2024},
  organization={IEEE}
}

@article{endonet,
author = {Andru Twinanda and Sherif Shehata and Didier Mutter and Jacques Marescaux and Michel De Mathelin and Nicolas Padoy},
year = {2016},
month = {02},
title = {EndoNet: A Deep Architecture for Recognition Tasks on Laparoscopic Videos},
volume = {36},
journal = {IEEE Transactions on Medical Imaging},
doi = {10.1109/TMI.2016.2593957}
}

\end{document}